\documentclass[letterpaper, 10 pt, conference]{ieeeconf}
\IEEEoverridecommandlockouts
\usepackage[dvipsnames]{xcolor}
\usepackage{verbatim}
\usepackage{amstext}
\usepackage{amsmath}

\usepackage{amsthm}
\usepackage{amssymb}
\usepackage{graphicx}
\usepackage{rotating}
\usepackage{todonotes}
\usepackage{csvsimple}

\usepackage{booktabs}
\let\labelindent\relax 

\usepackage[shortlabels]{enumitem}
\usepackage{algorithm}
\usepackage{algpseudocode}
\usepackage{setspace}
\usepackage{gensymb}

\makeatletter
\let\NAT@parse\undefined
\makeatother
\usepackage{hyperref}

\makeatletter
\let\NAT@parse\undefined
\makeatother

\newcommand{\algname}{STEADY}

\newcommand{\objective}{\Phi} 
\newcommand{\pair}[1]{\left\langle #1 \right\rangle}

\newcommand{\eg}{\textit{e.g.,}}

\title{\LARGE \bf
STEADY: Simultaneous State Estimation and Dynamics Learning from Indirect Observations
}

\author{Jiayi Wei$^{1}$, Jarrett Holtz$^{1}$, Isil Dillig$^{1}$, and Joydeep Biswas$^{1}$
\thanks{This work is supported in part by NSF (CCF-2019844), ARO (W911NF-21-20217), and a grant from the NSF Institute for Foundations of Machine Learning and the Machine Learning Laboratory at UT Austin.}
\thanks{$^{1}$Computer Science Department,
        University of Texas at Austin, USA.
        {\tt\small \{jiayi, jaholtz, isil, joydeepb\} @cs.utexas.edu}}%
}

\begin{document}

\maketitle
\thispagestyle{empty}
\pagestyle{empty}

\begin{abstract}
  Accurate kinodynamic models play a crucial role in many robotics applications such as off-road navigation and high-speed driving.
  Many state-of-the-art approaches for learning stochastic kinodynamic models, however, require precise measurements of robot states as labeled input/output examples, which can be hard to obtain in outdoor settings due to limited sensor capabilities and the absence of ground truth. 
  In this work, we propose a new technique for learning neural stochastic kinodynamic models from noisy and indirect observations by performing \emph{simultaneous state estimation and dynamics learning}. The proposed technique iteratively improves the kinodynamic model in an expectation-maximization loop, where the E Step samples posterior state trajectories using particle filtering, and the M Step updates the dynamics to be more consistent with the sampled trajectories via stochastic gradient ascent. 
  We evaluate our approach on both simulation and real-world benchmarks and compare it with several baseline techniques.
  Our approach not only achieves significantly higher accuracy but is also more robust to observation noise, thereby showing promise for boosting the performance of many other robotics applications.
\end{abstract}

\IEEEpeerreviewmaketitle

\section{Introduction}
\label{sec:intro}

Stochastic kinodynamic models are widely used in robotics applications such as state estimation, motion planning, and model-predictive control~\cite{thrun2002probabilistic}.  Such models are often written by hand and rely on large prediction uncertainty to compensate for model inaccuracy.  However, in many applications such as off-road navigation and high-speed robotics, more accurate models can lead to better downstream performance. For example, a learned neural kinodynamic model can better capture nonlinear effects like friction and saturation compared to a simple hand-written model. Prior work has proposed using supervised learning to learn ODEs for system dynamics~\cite{chen2018neural}, inertial and visual-inertial kinodynamic models~\cite{xiao2021learning, karnan2022viikd, atreya2022optimfkd}, and kinodynamic steering function~\cite{atreya2022s3f}. However, such supervised learning approaches require accurate measurements of robot states as labeled input/output data, which may not be available in many robotics applications due to limited sensing capabilities, such as when trying to learn the kinodynamic response of an off-road vehicle navigating at high speed in a dense forest.


In this work, we propose a new technique for learning stochastic kinodynamic models in settings where accurate state measurements are unavailable and we only have access to \emph{noisy, indirect} observations. At first glance, this appears to be a chicken-and-egg problem because learning the dynamics requires estimating the state trajectories from observations, whereas state estimation in turn requires access to an accurate kinodynamic model. We solve this apparent predicament by \emph{simultaneously} estimating the robot states while also learning the dynamics. Furthermore, our approach is data efficient and can be applied in settings where it is only feasible to collect training data from a few minutes of observation. 



\begin{figure}
  \begin{center}
  \includegraphics[width=0.85\linewidth]{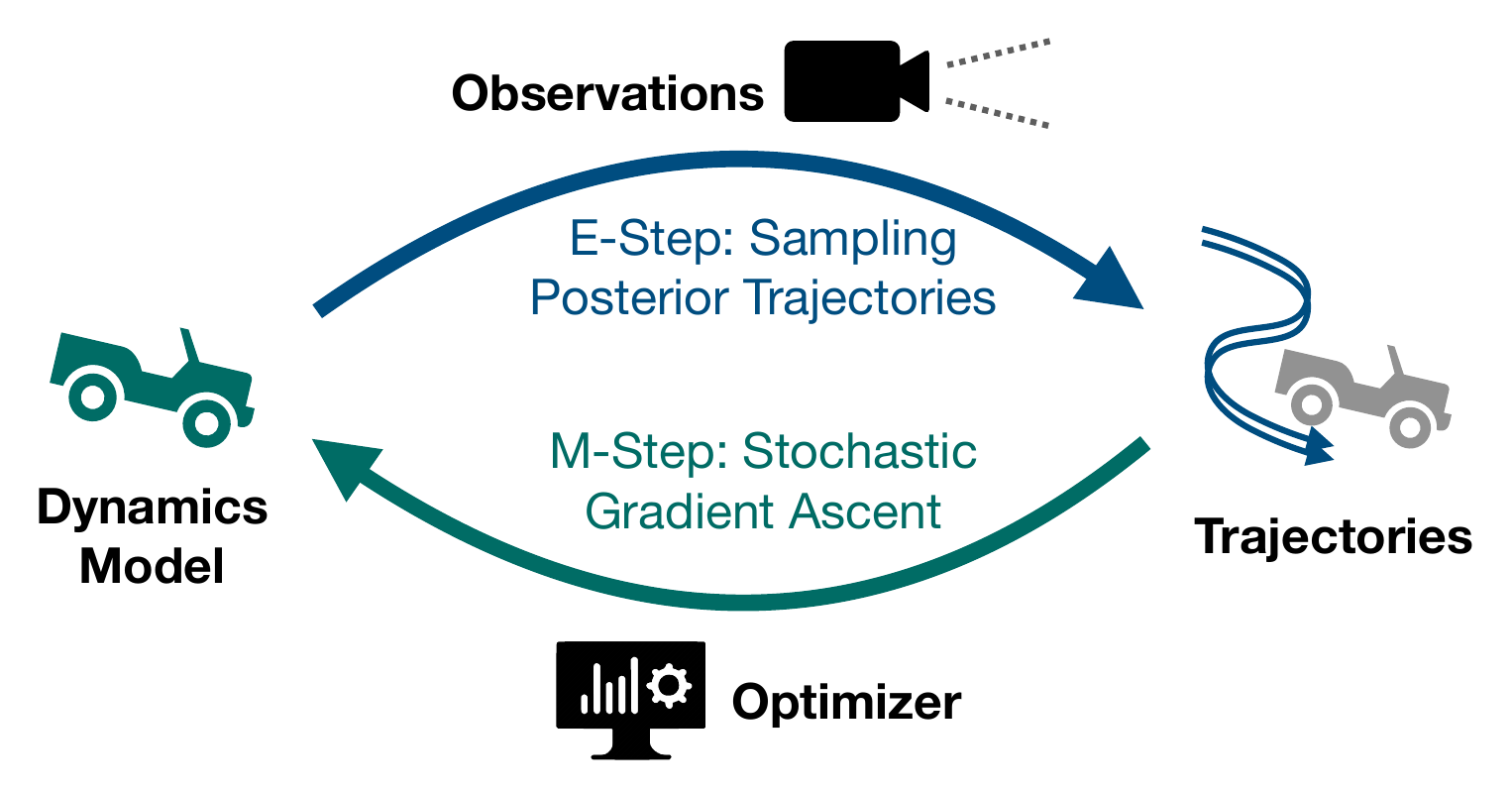}
  \caption{The high-level workflow of \algname.}
  \label{fig:workflow}
  \end{center}
  \vspace{-2em}
\end{figure}

At the heart of our approach lies an expectation-maximization (EM)~\cite{dempster1977maximum} loop that iteratively improves the neural kinodynamic model as well as our estimation of the robot states. 
Because EM methods work well for optimization problems with unobserved latent variables, this approach is a good fit for our setting wherein we cannot directly observe the robot states due to limited sensing capabilities. 
Given a fixed set of observations, our algorithm, \emph{simultaneous STate Estimation And DYnamics learning} (\algname{}), alternates between an E Step and an M Step, as illustrated in Figure~\ref{fig:workflow}. At a high-level, the E Step estimates a posterior probability density over trajectories by combining  both the current kinodynamic model and the imperfect sensor data. Then, the M Step updates the kinodynamic model using stochastic gradient ascent so that it becomes more consistent with the trajectories estimated using the E Step. 

Performing the E Step in our setting is challenging because the posterior is defined on a high-dimensional time-series distribution with no analytical solution, making the inference problem computationally intractable. To make matters worse, the E Step needs to be very fast because converging to a good model often requires thousands of EM iterations. To address these challenges,  \algname{}  uses a \emph{particle filtering} approach~\cite{doucet2009tutorial} that approximates the trajectory distribution in linear time and efficiently handles the underlying neural kinodynamic model representation.

Performing the M Step also requires special care because it can be very expensive to fully optimize the neural kinodynamic model in \emph{every} M Step, and the approximations made in the E Step also introduce Monte Carlo errors into the optimization objective. 
To mitigate these issues, we apply stochastic gradient ascent and only take a single gradient step in each M Step. We also apply a momentum-based optimizer like ADAM~\cite{kingma2014adam} to aggregate the gradient across multiple M Steps in order to dampen the effect of the Monte Carlo errors.

We evaluate our proposed algorithm on both simulated and real-world datasets. We compare \algname{} with several baseline approaches, including a recently developed learning technique based on stochastic variational inference and a hybrid approach that first performs state estimation and then applies supervised learning. Our main results show that \algname{} consistently outperforms  all  other  baselines and achieves performance that is similar to directly learning from the ground truth trajectories.

To summarize, we make the following contributions:
\begin{itemize}
  \item We introduce \emph{simultaneous state estimation and dynamics learning} as a new way to learn stochastic kinodynamic models for robotics applications and formulate the corresponding mathematical objective. 
  \item We present \algname{}, an expectation-maximization-style algorithm for solving the above problem  by combining particle filtering and stochastic gradient ascent.
  \item We provide a working, reusable implementation of \algname{} and experimentally demonstrate its superior effectiveness over other baseline approaches. 
\end{itemize}

\section{Problem Formulation} 
\label{sec:problem-formulation}

Given a system with state $x$ and control inputs $u$, we are interested in identifying the discrete-time forward kinodynamic function $f$ along with the zero-mean error distribution $Q_\epsilon$ such that the state dynamics is given by
\begin{align}
    x_{t+1} = f(x_{t},u_{t}) + \epsilon_t,\quad \epsilon_t \sim Q_\epsilon(\epsilon | x_t)\ .
\end{align}
Modeling $\pair{f, Q_\epsilon}$ is a challenging problem: while physical analysis can be used for simple kinematic systems such as skid-steer robots~\cite{rabiee2019friction}, the physics of vehicle to ground surface interactions (\eg{} for high-speed off-road driving) is poorly understood and must be learned directly from empirically observed data. Given a dataset $\pair{x_{1:T}, u_{1:T-1}}$ of a sequence of controls $u_{1:T-1}$ and the corresponding trajectory $x_{1:T}$, $\pair{f, Q_\epsilon}$ can be learned via supervised learning~\cite{xiao2021learning}. The trajectories in such supervised learning settings are either gathered using motion capture or using simultaneous localization and mapping (SLAM). 
However, collecting accurate trajectories $x_{1:T}$ in challenging settings is infeasible when motion capture is unavailable or when observations are sparse or noisy such that SLAM-estimated  trajectories are of insufficient accuracy for learning $\pair{f, Q_\epsilon}$.

Motivated by this problem, this work focuses on the more realistic setting where we cannot directly observe the states and have to instead rely on a \emph{noisy and indirect} observation sequence $y_{1:T} = (y_1, \ldots, y_T)$, which we assume are generated via some \emph{known} observation model
\begin{equation}
y_t \sim Q_y(y |\ x_t)\ , \label{eq:observation-model}
\end{equation}
where each observation $y_t$ carries only partial information about $x_t$ and can even be missing for certain time steps. As a result, each observation $y_t$ alone is generally not sufficient for uniquely determining the state $x_t$, and there will, in fact, be an infinite continuum of trajectories that are probabilistically compatible with the observation sequence $y_{1:T}$.

A common method for learning the dynamics under such noisy observations is to formulate it as a \emph{maximum a posteriori (MAP) estimation} problem, where the goal is to simultaneously find a \emph{single} trajectory $x_{1:T}$ and a dynamical model $\pair{f, Q_\epsilon}$ such that they maximize the posterior density
\begin{multline} 
    P(x_{1:T} | y_{1:T}, u_{1:T-1}) \propto P(x_{1:T}, y_{1:T} | u_{1:T-1})\\ =  
    \underbrace{
      Q_{x_1}(x_1) \prod_{t=1}^{T-1} Q_\epsilon(x_{t+1}-f(x_{t},u_{t})\ | x_t)
    }_{P(x_{1:T}|u_{1:T-1})} 
    \underbrace{
      \prod_{t=1}^T Q_y(y_t|x_t)
    }_{P(y_{1:T}|x_{1:T})}\label{eq:joint-density}
\end{multline}
where $Q_{x_1}$ is the distribution of the initial state.
However, this is an ill-formed objective for our setting since we are also  learning the error distribution $Q_\epsilon$.\footnote{
A simple way to see this is to consider the trivial solution $x_{t} \equiv 0$ and the corresponding (completely static) dynamics $f$. By reducing the variance of $Q_\epsilon$ toward zero, this solution allows us to drive the $Q_\epsilon$ term in \eqref{eq:joint-density} to an arbitrarily large value.
} To avoid this problem, we instead consider \emph{all} possible trajectories that are consistent with the observations (rather than a single trajectory) and find a dynamical model that maximizes the following \emph{marginalized} observation likelihood~\cite{bocquet2020bayesian}.
\begin{equation}
  P(y_{1:T} | u_{1:T-1}) = \int P(x_{1:T}, y_{1:T} | u_{1:T-1}) \,\mathrm{d}x_{1:T}\ . \label{eq:marginal-likelihood}
\end{equation}
This is a difficult optimization objective as it requires integrating the joint density over the trajectory space and has no analytical form for nonlinear dynamical systems. Prior works based on stochastic variational inference address this challenge by optimizing a tractable lower bound of \eqref{eq:marginal-likelihood}~\cite{krishnan2017structured, nguyen2020variational}. 
Such a lower bound is obtained by simultaneously training an \emph{inference network} in additional to the dynamical model. The inference network is trained to fit the posterior trajectory distribution $P(x_{1:T} | u_{1:T-1}, y_{1:T})$ and can thus predict the state trajectories from just observations and controls. However, as we will show in our evaluation, when the observation model $Q_y$ is complex, predicting the posterior distribution becomes a highly challenging task, and the inference network often fails to learn an adequate mapping, causing an inaccurate kinodynamic model to be learned. Thus, we explore an alternative approach that does not require training an inference network.

\section{The \algname{} Algorithm}

\begin{algorithm}[b]
\caption{The \algname{} Algorithm}\label{alg:overview}
 \textbf{Input:} control sequence $u_{1:T-1}$, observation sequence $y_{1:T}$, observation model $Q_y$, and initial state distribution $Q_{x_1}$. \\
 \textbf{Output:} $\theta$, the weights of the kinodynamic model $\pair{f, Q_\epsilon}$.
\begin{algorithmic}[1]\onehalfspacing
    \State Randomly initialize $\theta$ such that $f \approx \mathrm{identity}$ and $Q_\epsilon(\cdot)$ is sufficiently large.
    \While{$\pair{f, Q_\epsilon}$ has not overfitted}
    \State Run particle filtering forward in time using $Q_y$ and $Q_{x_1}$ to generate $N$ particles at each time step.
    \State Trace back $M$ trajectories  $\{x_{1:T}^i\}_{i=1}^M$ s.t. \eqref{eq:e Step-sampling} holds.
    \State Estimate $\tilde{\objective}(\theta)$ using the trajectories according to \eqref{eq:e Step-objective}.
    \State Take a gradient step w.r.t. $\theta$ to maximize $\tilde{\objective}(\theta)$.
    
    \EndWhile
\end{algorithmic}
\end{algorithm}


We now present our approach, \algname{}, for iteratively maximizing the optimization objective~\eqref{eq:marginal-likelihood}. 

\subsection{Algorithm Overview}
\algname{} is an instance of the \emph{generalized Monte Carlo expectation-maximization} algorithm~\cite{neal1998view}, where the E Step relies on a Monte-Carlo objective and the M Step makes incremental improvements to this objective. We summarize \algname{} in Algorithm~\ref{alg:overview} and elaborate on its key details in later subsections.

At a high level, \algname{} takes as input the control and observation sequence and outputs the learned kinodynamic model, $\langle f, Q_\epsilon \rangle$, which is represented as a neural network with parameters $\theta$. The parameters $\theta$ are initialized such that $f$ is approximately the identity function and $Q_\epsilon$ has large uncertainty (line 1 in Algorithm~\ref{alg:overview}). This initialization strategy ensures that the initial dynamics is stable and that it can be corrected from observation data. Then, the algorithm enters a loop (lines 2--6) in which the network parameters $\theta$ are iteratively updated to be more consistent with the observation. Specifically, lines 3-4 constitute the E Step of the algorithm and estimate a posterior probability density over trajectories through particle filtering (explained in more detail later). Then, the M Step in lines 5-6  uses the trajectories estimated in the E Step to update the neural network parameters $\theta$ by taking a gradient step.

In more detail, the E Step of \algname{} samples $M$ trajectories  $x^i_{1:T}$ from the posterior:
\begin{equation}
x_{1:T}^i \sim P(x_{1:T} |\ u_{1:T-1}, y_{1:T}),\quad i=1\ldots M\ .\label{eq:e Step-sampling}
\end{equation}
Note that samples $x^i_{1:T}$ are complete state trajectories. 
While we could, in principle, use factor graphs to approximate \eqref{eq:e Step-sampling} (by computing the full covariance around the MAP estimate of $x_{1:T}$), doing so would be both prohibitively expensive~\cite{kaess2009covariance} and also inaccurate when the posterior is multi-modal. Instead, we choose to estimate the posterior using particle filtering\footnote{
    Compared to more recently developed nonlinear smoothing techniques such as \emph{Forward Filtering/Back Simulation}~\cite{lindsten2013backward} or \emph{Particle Gibbs with Ancestor Sampling}~\cite{lindsten2014particle}, we use particle filtering since it has a better asymptotic time complexity and can be implemented efficiently when using the neural kinodynamic model.
}, as described in more detail in Section~\ref{sec:particle-filter}.



Given the trajectories obtained from the E Step, the M Step of \algname{} (lines 5-6 in Algorithm~\ref{alg:overview})  applies stochastic gradient ascent to adjust the model parameters $\theta$ in order to improve the following objective:
\begin{equation}
  \tilde{\objective}(\theta) = \frac{1}{M} \sum_{i=1}^M \log P_\theta(x^i_{1:T}, y_{1:T} | u_{1:T-1})\ , \label{eq:e Step-objective}
\end{equation}
where we write $P_\theta$ to emphasize that the model can affect $P_\theta(x^i_{1:T}, y_{1:T} | u_{1:T-1})$ via Eq.~\eqref{eq:joint-density}. To reduce the impact of sampling errors in the E Step, we do not run this optimization to convergence but instead only take a single gradient step. This design choice also reduces the computational overhead of training the neural network in each M Step.

\subsection{Improvement Guarantees}
We now briefly explain why our proposed algorithm maximizes the objective~\eqref{eq:marginal-likelihood}, borrowing standard assumptions/results from the Monte Carlo expectation-maximization literature (\eg{} see \cite{neal1998view}). 

Denote the value of $\theta$ in the E Step as $\theta_E$. Since $x^i_{1:T}$ are sampled from $P_{\theta_E}(x^i_{1:T}|y_{1:T})$, we can view the M Step objective $\tilde{\objective}(\theta)$ in \eqref{eq:e Step-objective} as the Monte Carlo estimate of another objective $\objective(\theta, \theta_E)$, defined as
\begin{multline}
\objective(\theta, \theta_E) = \int \log P_\theta(x_{1:T}, y_{1:T})     P_{\theta_E}(x_{1:T}|y_{1:T}) \mathrm{d} x_{1:T} \\
 = \mathop{\mathbb{E}}_{x_{1:T}^i} \left[\log P_\theta(x^i_{1:T}, y_{1:T} | u_{1:T-1})\right] \approx
    \tilde{\objective}(\theta) \label{eq:ideal-objective}
\end{multline}

Taking the difference between $\objective(\theta, \theta_E)$ and $\objective(\theta_E, \theta_E)$ , we then have
\begin{multline}
  \objective(\theta, \theta_E) - \objective(\theta_E, \theta_E) \\
  = \int \log \frac{P_\theta(x_{1:T} | y_{1:T}) P_\theta(y_{1:T})}{P_{\theta_E}(x_{1:T}| y_{1:T}) P_{\theta_E}(y_{1:T})} P_{\theta_E}(x_{1:T}|y_{1:T}) \mathrm{d}x \\ 
  = \underbrace{\int \log \frac{P_\theta(x_{1:T} | y_{1:T})}{P_{\theta_E}(x_{1:T} |  y_{1:T})} P_{\theta_E}(x_{1:T}|y_{1:T}) \mathrm{d}x}_{\text{(negative KL divergence)}} \\  + \log P_\theta(y_{1:T}) - \log P_{\theta_E}(y_{1:T})\\
  \leq \log P_\theta(y_{1:T}) - \log P_{\theta_E}(y_{1:T})\ .\label{eq:improvement}
\end{multline}
This shows that any improvement  on $\objective(\theta, \theta_E)$ will be a lower bound of the improvement on $\log P_\theta(y_{1:T})$. Thus, when our Monte Carlo estimate $\tilde{\objective}(\theta)$ closely approximates $\objective(\theta, \theta_E)$\footnote{
To ensure that the Monte Carlo estimate is a good approximation of $\objective(\theta, \theta_E)$, recall that \algname{} takes a single optimization step using a momentum-based optimizer, which ensures that most Monte Carlo errors can cancel out across  M Steps.  
}, optimizing $\tilde{\objective}(\theta)$ also optimizes $P_\theta(y_{1:T})$, which is the objective stated in \eqref{eq:marginal-likelihood}. 

\subsection{Implementation}

In this section, we describe the key design choices behind our implementation of the \algname{} algorithm. 

\subsubsection{Kinodynamic Model Architecture} \label{sec:nn-architecture}

\begin{figure}
  \begin{center}
  \includegraphics[width=0.7\linewidth]{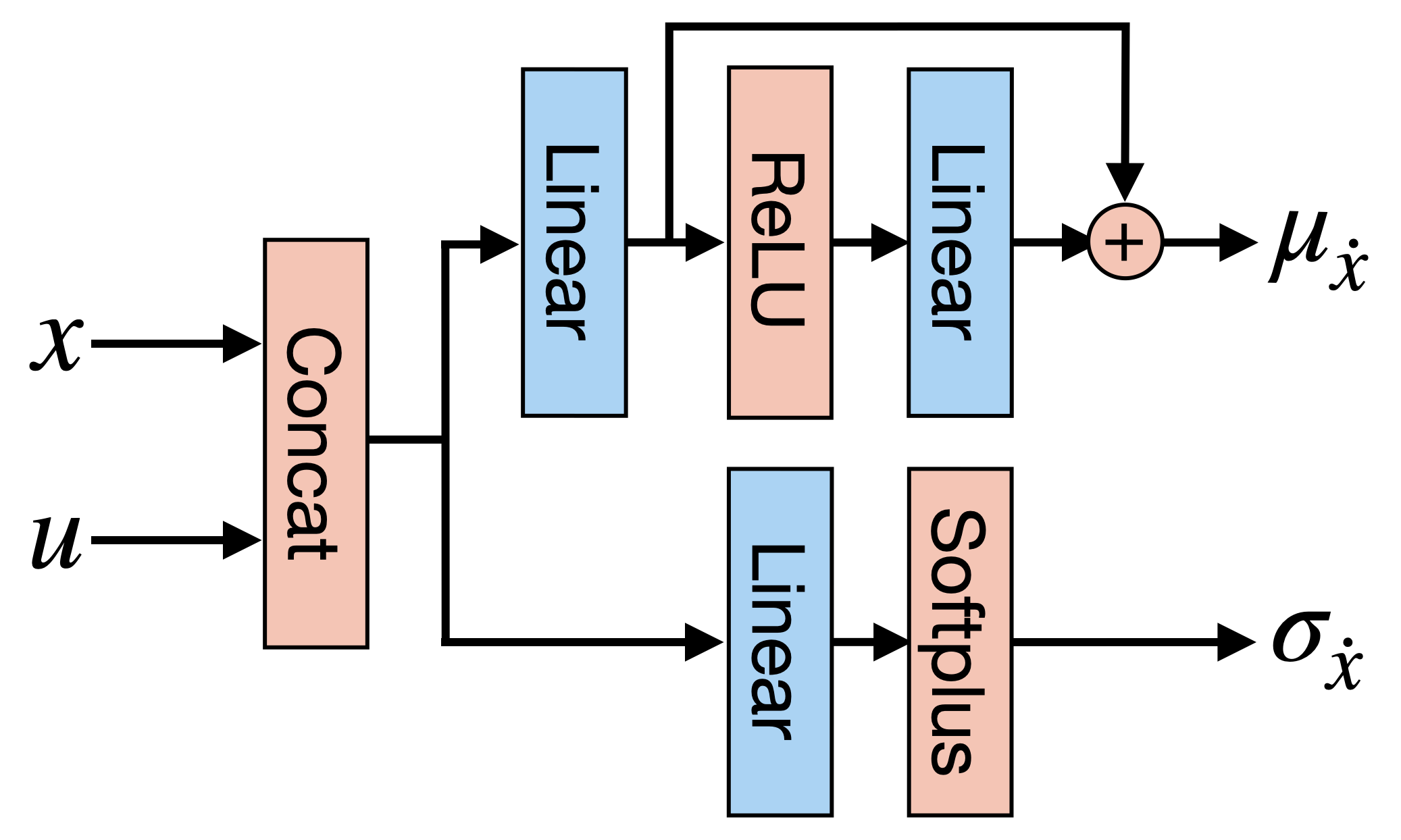}
  \caption{Architecture of the dynamics network. $x$ and $u$ are the current state and control, and $\mu_{\dot{x}}$ and $\sigma_{\dot{x}}$ are the mean and variance of the predicted state derivatives. All inputs/outputs are defined in local reference frames.}
  \label{fig:dynamics-net}
  \end{center}
\end{figure}

In our implementation, we represent the kinodynamic model $\pair{f, Q_\epsilon}$ using the neural architecture shown in Figure~\ref{fig:dynamics-net}. This neural network takes as input the current state and control in the local reference frame and predicts the state derivative $\dot{x}$ (in the same reference frame) using a diagonal normal distribution with mean $\mu_{\dot{x}}$ and variance $\sigma_{\dot{x}}$. We compute the mean $\mu_{\dot{x}}$ using a fully connected network with 64 ReLU activation units and one hidden layer. We also add a skip connection to make it easier for the network to learn the linear part of the dynamics~\cite{he2016deep}. We compute the diagonal elements of the covariance matrix using a simple linear layer followed by the softplus activation $\mathrm{softplus}(x) = \log(1 + \exp(x))$ to ensure that the outputs are always positive.

\subsubsection{Sampling Posterior Trajectories with Particle Filtering}
\label{sec:particle-filter}

To sample posterior trajectories using a particle filter, we first run the particle filter forward in time to obtain a fixed number of particles for each time step. This process involves alternating between the prediction and resampling step. In each prediction step, new particles for the next time step are sampled from $P(x_{t+1}|x_t, u_t)$ by forward-simulating the dynamical model. In each resampling step, we use the observation model $P(y_{t+1}|x_{t+1})$ to replicate those particles that are more consistent with the new observation $y_{t+1}$ and discard those that are less so. Once we have obtained the particles at the last time step, we randomly pick one of them and trace its ancestral lineage back to the first time step to obtain a complete state trajectory.\footnote{
Although particle filtering is prone to the path degeneration issue (where most trajectories share the same ancestor due to excessive resampling),
this issue tends to get better as the dynamical model becomes more consistent with the observations as training progresses. Empirically, we only observe severe path degeneration in early stages of training, at which time even a very noisy gradient of $\tilde{\objective}(\theta)$ is sufficient to push the parameters toward the right direction.
}  When the number of particles is sufficiently large, this approach gives unbiased samples from the posterior distribution~\cite{doucet2009tutorial}. In each E Step, we run the particle filter once to sample $N$ particles at each time step and reuse the particles to sample $M$ trajectories. To support efficient computation with the neural network, we store all particles using the structure-of-arrays pattern and perform all operations in batch.

\subsubsection{Accelerating Learning by Flattening the E Step posterior} \label{sec:flatten-e-step}
Since the kinodynamic model is very inaccurate in the early stages of  training, we found that directly using the E Step posterior $P(x_{1:T}|u_{1:T-1}, y_{1:T})$ can cause all sampled trajectories to concentrate around a single incorrect trajectory, slowing down the learning process or even trapping the optimizer in local optima. To mitigate this problem, we take inspiration from simulated annealing and modify the E Step to instead sample with a modified observation model
\begin{equation*}
    y_t \sim Q_y(y |\ x_t)^{w_{\mathrm{obs}}}\ ,
\end{equation*}
where the parameter $w_{\mathrm{obs}} \in [0, 1]$ controls the importance of the observations. During the course of the training, we linearly increase $w_{\mathrm{obs}}$ from 0 to 1. As shown in our evaluation in Section~\ref{sec:obs_noise_impact}, this modification consistently improves the training outcome.

\subsubsection{Code availability}
Our implementation is written in Julia~\cite{Julia-2017} and is built on top of the Flux machine learning library~\cite{Flux.jl-2018}. We publish our code on Github.\footnote{Repository: \texttt{\url{https://github.com/MrVPlusOne/STEADY}}}

\section{Evaluation}
In this section, we first describe our experimental setup and then present the results to answer: (1) Can \algname{}  achieve better performance  compared to relevant baselines?  (2) How do various factors and design choices affect the performance of \algname{}?


\subsection{Experimental Setup}

Before presenting our  empirical results, we first discuss the benchmarks and baselines used in the evaluation as well as other relevant experimental details.

\subsubsection{Benchmarks} We perform our evaluation on the following datasets, which consist of both simulated and real-world data:
\begin{itemize}[leftmargin=*]
  \item \textbf{Hov} is a dataset that we create by simulating a hovercraft moving on a 2D plane with nonlinear air resistance. We control the hovercraft by adjusting the thrust of its propellers and simulate 16, 32, and 32 trajectories for training, validation, and testing, respectively. Each trajectory is 10 seconds long.
  \item \textbf{Hov+} is a data-rich variant of Hov that contains 10 times more training trajectories. 
  \item \textbf{Truck} is a real-world dataset obtained by tele-operating a scale 1/5, four-wheel drive, Ackermann steering vehicle that we call the AlphaTruck. We divide the collected data into 10-second trajectories and use about 2 minutes of data each for training, validation, and testing.
  
\end{itemize}

For all three datasets, we represent the robot state using 6 variables (i.e, 3 variables for the 2D pose and the other 3 for the velocities). 
We use simulated observations to control the effects of observation noise. We generate these observations from the ground-truth states using a landmark-based, bearing-only observation model. For each benchmark, we randomly place 4 landmarks and measured their relative angles w.r.t. the robot's pose at every time step. We also add a small amount of Gaussian noise ($\sigma=5\degree$) to the measured angles to simulate sensor noise.

\subsubsection{Baseline approaches} \label{sec:baselines}
We compare \algname{} with the following baselines:
\begin{itemize}[leftmargin=*]
  \item \textbf{Hand:} This baseline involves hand-written kinodynamic models customized for each of the three datasets. Specifically, for
 the Hov datasets, the hand-written model is the ground-truth dynamics with two modifications: (1) it does account for  air friction; (2) it compensates for model inaccuracy by using a disturbance variance that is twice the ground truth. For the Truck dataset, the hand-written model is the classical kinematic model that assumes bounded forward acceleration and no sliding along the lateral direction.

  \item \textbf{FitHand:}  This baseline first performs state estimation using  particle filtering with the hand-written model. It then applies supervised learning to fit an updated model to the estimated trajectories. This method can be viewed as a simplified version of \algname{} that performs both state estimation and learning but without embedding them in the EM loop. 
  
  \item \textbf{FitTV:} This baseline performs supervised learning without relying on \emph{any} motion model. In particular, it first estimates the most likely pose at each time step using only the observation model and then applies total variation regularization~\cite{chartrand2011numerical, brunton2016discovering} to estimate the (de-noised) velocities. This baseline is relevant for exploring the impact of an explicit motion model on performing state estimation. 
  \item \textbf{SVI:}  This baseline learns a kinodynamic model using stochastic variational inference, as described in  \cite{krishnan2017structured}. To adapt this approach to our setting, we replace the GRU-like dynamics network proposed in  \cite{krishnan2017structured}  with the neural architecture  described in Section~\ref{sec:nn-architecture}.
  \item \textbf{FitTruth:} As an upper bound on the learning performance, we also report the results of applying supervised learning directly on  ground truth states.
\end{itemize}


\subsubsection{Hyperparameters} 
In the following experiments, we use at most 40,000 EM steps, leveraging the validation set to early-stop the training (according to the estimated marginal observation density $P(y_{1:T})$).
In our E Step implementation, we use $N=20,000$ particles and sample $M=10$ trajectories. When the training data contains multiple trajectories, we only use the subset from a single trajectory in each E Step.
In our M Step implementation, we use the ADAM optimizer with a fixed learning rate of $10^{-4}$ and all other parameters set to their default values. We also normalize the loss  by dividing it by the length of the trajectory.

\subsection{Main Results}

In this section, we report the results of our empirical comparison between \algname{} and the five baselines described earlier. Specifically,  we evaluate the (learned) model's performance in terms of state estimation (Table~\ref{tab:state-estimation-compare}) and forward prediction error (Table~\ref{tab:forward-baselines}). Both metrics measure the root-mean-square error (RMSE) between the poses obtained by the  model and the ground truth.\footnote{
We compute the state estimation error using the posterior trajectories obtained from particle filtering and  reduce the observation frequency from 10Hz (at training time) to 1Hz (at test time) to make the effect of the kinodynamic model more salient.
We compute the forward prediction error by taking each state from the ground truth trajectories as the initial state, running the learned model forward for 10 time steps (using the recorded controls only), and measuring the RMSE between the predicted final pose and the ground truth.
} We run each baseline 3 times and report the best performance on the test set.

\newcommand{\tabheader}[2]{\multicolumn{1}{#1}{#2}}

\begin{table}
\caption{\label{tab:state-estimation-compare} State Estimation Performance Comparisons}
\centering
\small
\begin{tabular}{l|lll|lll}
    \toprule
    & \multicolumn{3}{c|}{\textbf{Location Error (m)}} & \multicolumn{3}{c}{\textbf{Angular Error (rad)}} \\
    & \tabheader{c}{Hov} & \tabheader{c}{Hov+} & \tabheader{c|}{Truck} & \tabheader{c}{Hov} & \tabheader{c}{Hov+} & \tabheader{c}{Truck}\\
    \midrule
    Hand & 1.168 & 1.145 & 0.554 & 0.574 & 0.532 & 0.103   \\
    FitHand & 0.505 & 0.491 & 0.446 & 0.040 & 0.037 & 0.087 \\
    FitTV & 1.381 & 1.355 & 0.504 & 0.144 & 0.092 & 0.108 \\
    SVI & 0.637 & 0.366 & 0.432 & 0.050 & 0.038 & 0.090 \\
    \algname{} & \textbf{0.375} & \textbf{0.346} & \textbf{0.258} & \textbf{0.033} & \textbf{0.031} & \textbf{0.063} \\
    \midrule
    \emph{FitTruth} & \emph{0.368} & \emph{0.314} & \emph{0.238} & 
    \emph{0.035} & \emph{0.030} & \emph{0.063}\\
    \bottomrule
\end{tabular}
\end{table}

\begin{table}
\caption{\label{tab:forward-baselines} Forward Prediction Performance Comparisons}
\centering
\small
\begin{tabular}{l|lll|lll}
    \toprule
    & \multicolumn{3}{c|}{\textbf{Location Error (m)}} & \multicolumn{3}{c}{\textbf{Angular Error (rad)}} \\
    & \tabheader{c}{Hov} & \tabheader{c}{Hov+} & \tabheader{c|}{Truck} & \tabheader{c}{Hov} & \tabheader{c}{Hov+} & \tabheader{c}{Truck}\\
    \midrule
    Hand & 0.058 & 0.059 & 0.472 & 0.025 & 0.024 & 0.151  \\
    FitHand & 0.050 & 0.051 & 0.373 & 0.033 & 0.030 & 0.107 \\
    FitTV & 0.131 & 0.147 & 0.461 & 0.089 & 0.087 & 0.213 \\
    SVI & 0.088 & 0.046 & 0.229 & 0.036 & 0.025 & 0.080 \\
    \algname{} & \textbf{0.042} & \textbf{0.038} & \textbf{0.200} & \textbf{0.013} & \textbf{0.013} & \textbf{0.079} \\
    \midrule
    \emph{FitTruth} &  \emph{0.031} & \emph{0.027} & \emph{0.174} & \emph{0.017} & \emph{0.012} & \emph{0.085} \\
    \bottomrule
\end{tabular}
\end{table}

As shown in Table~\ref{tab:state-estimation-compare} and \ref{tab:forward-baselines}, \algname{} consistently outperforms all other baselines (excluding FitTruth) on all datasets under both metrics. Furthermore, \algname{} in many cases achieves a performance level that is very similar to FitTruth. Comparing the performance of Hand and FitHand, we see that learning from the data generated by a  hand-written model can lead to performance improvement; however, the performance of FitHand is still considerably worse than that of \algname{}. Comparing FitTV and FitHand, we observe that FitHand performs better than FitTV, which suggests that learning with an inaccurate motion model is still better than without one. Finally, the SVI approach is more data-hungry and only gets close to \algname{}'s performance on the data-rich Hov+ dataset. 

\begin{figure*}
\begin{center}
  \includegraphics[width=0.8\textwidth]{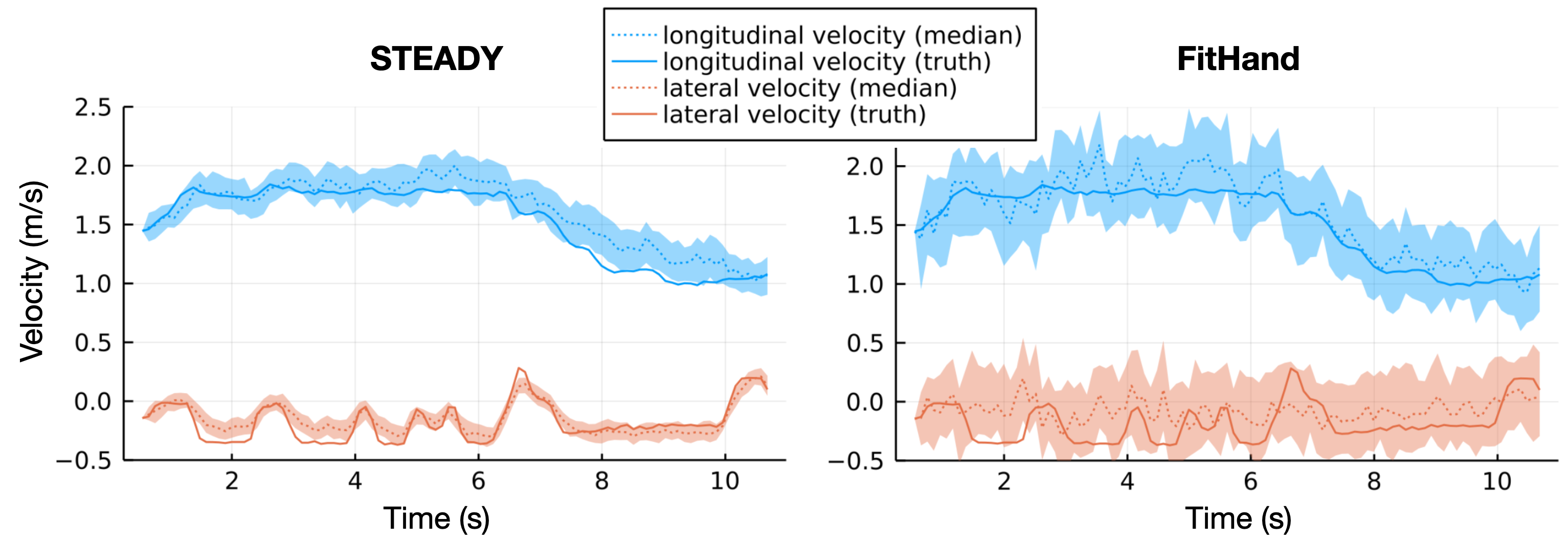}
  \caption{\label{fig:posterior-visualization} Posterior velocity visualization. We show the posterior quality of \algname{} (left) and FitHand (right) on one test trajectory from the Truck dataset. Both the longitudinal and lateral velocities are measured in the robot's local reference.}
\end{center}
\vspace{-0.7cm}
\end{figure*}

To qualitatively validate our results, we also visualize the posterior velocities in Figure~\ref{fig:posterior-visualization} for FitHand and \algname{}. As shown in this figure,  \algname{} is able to closely track the overall trends in the lateral motion (show in orange), whereas FitHand's lateral prediction is much more uncertain and resembles  white noise. This is consistent with our intuition that using a flawed hand-written model (which assumes no lateral sliding) can introduce biases to the training data and subsequently lead to worse performance.



\begin{figure}
  \begin{center}
    \includegraphics[width=0.9\linewidth]{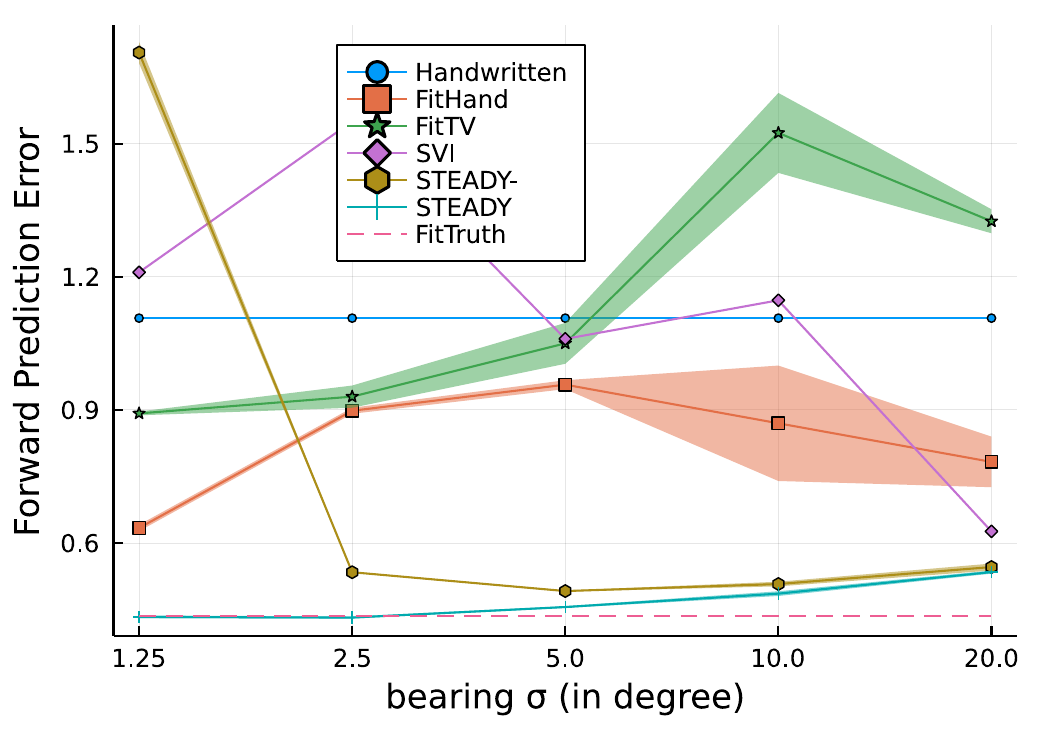}
  \end{center}
  \caption{\label{fig:vary_obs_noise} How observation noise affects forward prediction error. All learning-based approaches are trained on the Truck dataset using the bearing-only observation model. }
\end{figure}

\subsection{Impact of Observation Noise} \label{sec:obs_noise_impact}

In this section, we perform an evaluation to assess the impact of  observation noise on both \algname{} and the baselines. Additionally, we also consider an ablation of \algname{} called \algname{}- that does not utilize the E Step flattening modification proposed in Section~\ref{sec:flatten-e-step}.  To perform this evaluation, we vary the magnitude of
the angular noise in the Truck dataset between 1.25 and 20 degrees. We run each approach 3 times (except for SVI, which we only run once due to its much longer training time) and report the forward prediction error in Figure~\ref{fig:vary_obs_noise}. 
Overall, these results follow the expected relationship
between noise and model performance: in general, all approaches (except Hand and FitTruth, which do not use the observations for training) tend to get worse as the magnitude of noise increases. However, the performance impact on \algname{} is relatively small, demonstrating its robustness to observation noise, and it is also the only approach whose performance converges to that of FitTruth on low-noise settings. We can also see that \algname{}- performs significantly worse than \algname{} when the noise is low, suggesting that the E Step flattening modification is crucial for stable training under a sharp posterior.

\begin{figure}
  \begin{center}
    \includegraphics[width=0.9\linewidth]{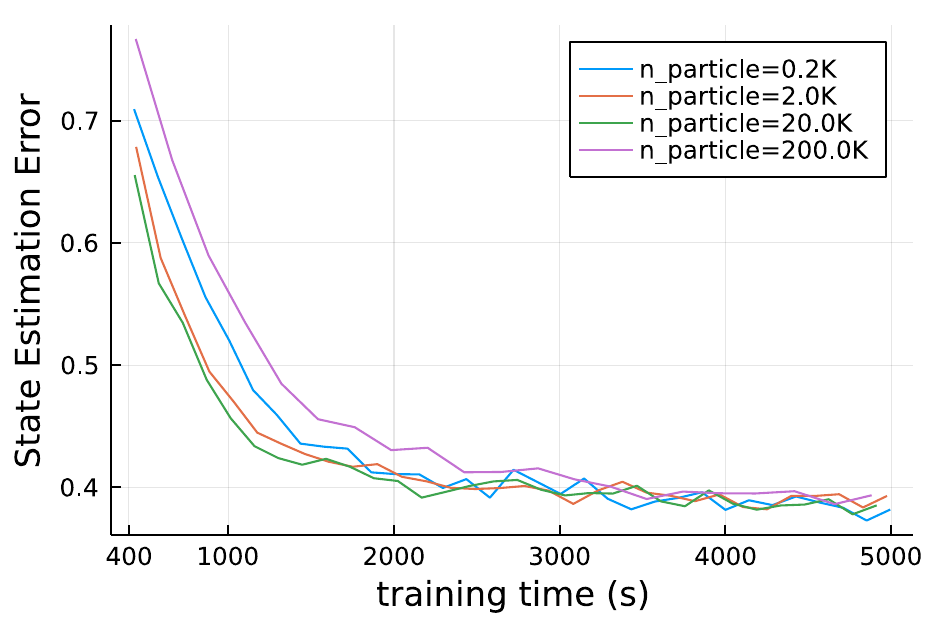}
  \end{center}
  \caption{\label{fig:vary_particles} How the number of particles in the E Step affects the learning speed and accuracy of \algname{}. Using 20K particles leads to fastest training in terms of running time, but all four settings eventually converge to similar levels of performance.}
\end{figure}

\subsection{Impact of Particle Quantity}

In this section, we report the results of an experiment that is designed to evaluate the impact of the particles used in the E Step, which is one of the most important hyper-parameters underlying the \algname{} implementation. To this end, we vary the number of
particles used for the Truck dataset from $200$ to $200,000$ and plot the validation set performance against training time in Figure~\ref{fig:vary_particles}.

The results of this experiment demonstrate the known
trade-off between state estimation accuracy and computational cost. With more particles, the estimated gradient in the M Step has higher quality and hence leads to faster learning in terms of training steps. However, more particles also incur a longer running time. Overall, we find that the sweet spot in our setting lies somewhere between $2000$ and $20,000$ particles. Another interesting finding from this experiment is that, no matter how many particles are used, all configurations eventually converge to a similar performance before starting to overfit, suggesting that overfitting might be the dominating factor here (compared to Monte Carlo error).

\section{Related Work}

In this section, we survey  robotics literature  on learning dynamical models and then review techniques for learning stochastic dynamical systems in a more general context.

\noindent
\textbf{Dynamics Learning in Robotics. }
Most recent publications on robot dynamics learning target different settings than the one we consider here. For example, \cite{xiao2021learning} aims to learn inverse kinematic models in a deterministic setting and assumes the training data consist of directly observed, accurate robot states, whereas we focus on learning forward dynamical models from noisy and indirect observations. \cite{li2019propagation} focuses on learning the interaction between multiple objects and also assumes the underlying dynamics are deterministic. That work also differs from ours in how the state space is represented: while their learned dynamics operate in opaque, high-dimensional latent spaces, ours are defined in terms of low-dimensional, physically grounded states. An alternative method to handle partial state observability is by applying the Koopman Operator Theory, as done in \cite{bruder2019nonlinear} to learn deterministic dynamical models of soft-body robots. However, this approach is only applicable when there is little noise in the observations. Finally, \cite{fang2019dynamics} learns stochastic dynamical models for hierarchical planning of manipulation tasks, and, similar to \cite{li2019propagation}, it also performs learning in a high-dimensional latent state space.

\noindent
\textbf{Learning Stochastic Dynamical Systems. }
Outside the field of robotics, there is a rich literature on data-driven identification of stochastic dynamical systems. Prior approaches can be divided into two main categories: those based on expectation-maximization (EM) v.s. those based on stochastic variational inference (SVI). The EM category includes the pioneer work of \cite{ghahramani1998learning}, which parameterizes the dynamics as radial basis function approximators and uses Extended Kalman Smoothing for the E Step. More recent work includes \cite{schon2011system}, which implements the E Step using particle smoothing, and \cite{bocquet2020bayesian}, which focuses on learning partial differential equations to model the geophysical flow dynamics. As for the SVI category, one representative approach is the one proposed in \cite{krishnan2017structured}, upon which our SVI baseline is  based. More recent work \cite{nguyen2020variational} also applies SVI to learning chaotic dynamical systems in  more data-rich settings.

\section{Conclusion and Future Work} 
\label{sec:conclusion}

We introduced the simultaneous state estimation and dynamics learning problem for learning kinodynamic models from noisy and indirect data. Our key idea is to iteratively improve both the state estimation and the learned kinodynamic model using the expectation-maximization framework. 
Our approach is simple to implement, works for real-world robots, and outperforms existing approaches.

In the future, we plan to extend our approach to more challenging settings such as simultaneous SLAM and dynamics learning. We also plan to explore ways to learn interpretable dynamical models by combining ideas from program synthesis and symbolic regression.



\bibliography{references} 
\bibliographystyle{ieeetr}

\end{document}